\def\year{2022}\relax
\documentclass[letterpaper]{article} % DO NOT CHANGE THIS
\usepackage{aaai22}  % DO NOT CHANGE THIS
\usepackage{times}  % DO NOT CHANGE THIS
\usepackage{helvet}  % DO NOT CHANGE THIS
\usepackage{courier}  % DO NOT CHANGE THIS
\usepackage[hyphens]{url}  % DO NOT CHANGE THIS
\usepackage{graphicx} % DO NOT CHANGE THIS
\usepackage{natbib}  % DO NOT CHANGE THIS AND DO NOT ADD ANY OPTIONS TO IT
\usepackage{caption} % DO NOT CHANGE THIS AND DO NOT ADD ANY OPTIONS TO IT
\DeclareCaptionStyle{ruled}{labelfont=normalfont,labelsep=colon,strut=off} % DO NOT CHANGE THIS
\usepackage{algorithm}
\usepackage{algorithmic}
\usepackage{booktabs}
\usepackage{subcaption}
\usepackage{tikz, pgfplots}

\usepackage{newfloat}
\usepackage{listings}
\floatstyle{ruled}
\newfloat{listing}{tb}{lst}{}
\floatname{listing}{Listing}
\nocopyright
\title{Handling Heavy Occlusion in Dense Crowd Tracking \\by Focusing on the Heads}
\author{
    Yu Zhang,
    Huaming Chen,
    Wei Bao,
    Zhongzheng Lai,
    Zao Zhang,
    Dong Yuan,
}
\affiliations{
    Univerisy of Sydney\\
    yu.zhang1, huaming.chen, wei.bao zhongzheng.lai, zao.zhang, dong.yuan@sydney.edu.au
}

\usepackage{bibentry}
\begin{document}

\maketitle
\begin{abstract}
% AAAI creates proceedings, working notes, and technical reports directly from electronic source furnished by the authors. To ensure that all papers in the publication have a uniform appearance, authors must adhere to the following instructions.

With the rapid development of deep learning, object detection and tracking play a vital role in today's society. Being able to identify and track all the pedestrians in the dense crowd scene with computer vision approaches is a typical challenge in this field, also known as the Multiple Object Tracking (MOT) challenge. Modern trackers are required to operate on more and more complicated scenes. According to the MOT20 challenge 
result, the pedestrian is 4 times denser than the MOT17 challenge. Hence, improving the ability to detect and track in extremely crowded scenes is the aim of this work. In light of the occlusion issue with the human body, the heads are usually easier to identify. In this work, we have designed a joint head and body detector in an anchor-free style to boost the detection recall and precision performance of pedestrians in both small and medium sizes. Innovatively, our model does not require information on the statistical head-body ratio for common pedestrians detection for training. Instead, the proposed model learns the ratio dynamically. To verify the effectiveness of the proposed model, we evaluate the model with extensive experiments on different datasets, including MOT20, Crowdhuman, and HT21 datasets. As a result, our proposed method significantly improves both the recall and precision rate on small\&medium sized pedestrians and achieves state-of-the-art results in these challenging datasets.
\end{abstract}

\section{Introduction}

Tracking by detection is one of the most critical framework for the Multiple Object Tracking (MOT) challenge. Since the modern MOT challenge is now prone to happen in more and more crowded scenes, the tracker performance is limited by the design of the object detectors. With a latest detector, it is possible to achieve a near state-of-the-art performance in terms of tracking accuracy with a traditional tracking algorithm. In the meantime, many approaches with novel architectures have been proposed for the pedestrian detection task. Earlier works like Faster-RCNN \cite{ren2015faster}, Mask-RCNN \cite{he2017mask}, SSD \cite{liu2016ssd} utilize anchor-based model framework while recently researches about anchor-free detection have started to attract more and more attention \cite{zhang2020bridging}.

The key challenge of the pedestrian detection is occlusion in the scene. In crowded scenes, people overlap with each other most of the times. The challenge of occlusion makes it hard for the detection model to extract useful features to identify the person in an effective way. It also cause problem with the Re-ID functions with some models because the visual features required to identify the person constantly changes due to occlusion. Multiple previous researches have stated that exploiting multiple part of a pedestrian helps to improve the performance of pedestrian detection \cite{chi2020relational, zhou2018bi, chi2020pedhunter}.

% \begin{figure}[t]
% \begin{center}
% \includegraphics[width=0.9\linewidth]{fig/teaser.pdf}
   
% \end{center}
%    \caption{MOTA-IDF1 comparisons of different trackers on the test set of MOT20. With the assist of our supplementary body detector, the proposed tracking framework achieves 78.2 MOTA and 75.5 IDF1 score, outperforming all previous trackers.}
% \label{fig:teaser}
% \end{figure}
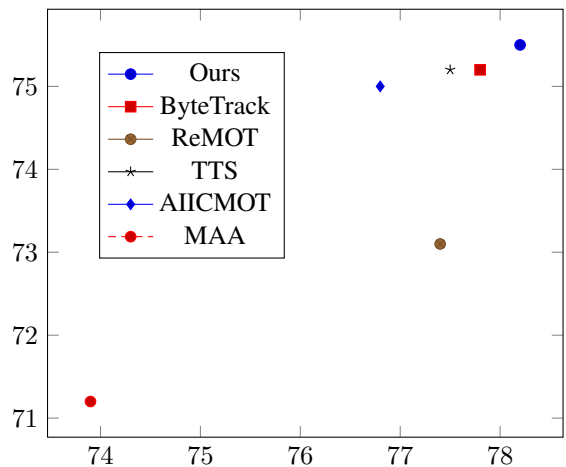
\begin{figure}[t]
\centering
\begin{tikzpicture}
\begin{axis}[legend style={at={(0.1,0.9)},anchor=north west}] 
\addplot 
table                              
{           		                % X，Y raw data
 X Y
 78.2 75.5
};
\addplot
table
{   				
 X Y
 77.8 75.2
};
\addplot 
table                              
{           		                % X，Y raw data
 X Y
 77.4 73.1
};
\addplot 
table                              
{           		                % X，Y raw data
 X Y
 77.5 75.2
};
\addplot 
table                              
{           		                % X，Y raw data
 X Y
 76.8 75.0
};
\addplot 
table                              
{           		                % X，Y raw data
 X Y
 73.9 71.2
};
\addlegendentry{Ours}        
\addlegendentry{ByteTrack}
\addlegendentry{ReMOT}
\addlegendentry{TTS}
\addlegendentry{AIICMOT}
\addlegendentry{MAA}
\end{axis}

\end{tikzpicture}
\caption{MOTA-IDF1 comparisons of different trackers on the test set of MOT20. With the assist of our supplementary body detector, the proposed tracking framework achieves 78.2 MOTA and 75.5 IDF1 score, outperforming all previous trackers.}
\label{fig:teaser}
\end{figure}

According to previous research \cite{chi2020relational}, human body detection are often missing due to the large overlaps in extremely crowded scenes. The model either failed to identify enough features for the occluded person or eliminate them during the NonMaximum Suppression (NMS) post-process procedure. Compared to body detection, the occlusion between heads are less likely to occur. With this consideration, the feature map can become more distinguishable and consistent for training. By comparing the head detection results and body detection results on the same MOT challenge sequence, we find out that the head detector generally identifies more people than the body detector (except for those people not showing full bodies in the scene). Recently, \cite{sundararaman2021tracking} introduced a new head tracking dataset to further assist tracking humans efficiently in densely crowded environments. Hence, by combining head tracking with body tracking detectors, we should be able to improve the performance of the MOT challenge. 

However, adopting head tracking to body tracking is not a trivial task. First of all, many pedestrians do not show their heads in the camera sight. Head detector alone cannot fully replace the body detector in the context of MOT. Also, in a traditional detector, human heads are detected separately. Manually generating body bounding boxes from head detection introduce lots of false positive as people have different poses. Hence, detecting both head and body simultaneously then linking them together is challenging. Bi-Box \cite{zhou2018bi} proposed a model to estimate both visible part and the full body at the same time. JointDet \cite{chi2020relational} and HBAN\cite{lu2020semantic} are two anchor-based solution which rely on a fixed head-body ratio to generate the head proposal from the regular body proposal. The problem with these methods is that, they still rely on anchors or external method to generate head and body proposals, leading to low recall rate and inaccurate body Region of Interest (RoI). Although most standing human follows a static head-body ratio, it still varies a lot due to different human poses and camera positions. Using a fixed head-body proposal to predict pedestrians in complex scenes leads to a less satisfying result. 

Motivated by the findings, we propose JointTrack as a novel solution. The contributions of this work are three folds: 1). We adopt the joint head and body detector in an anchor-free style to further boost its detection performance in extremely crowded scenes, and a supplementary body detector is leveraged to overcome its limitation in the MOT challenge; 2). Instead of relying on a fixed head-body ratio, a new module to include both the head and body prediction based on SimOTA is proposed, which learns the relationship dynamically during training; 3). the proposed model achieves state-of-the-art head detection performance on CrowdHuman dataset, and the joint head and body tracking framework achieves state-of-the-art performance in the MOT20, HT21 challenges.

\begin{figure*}
\begin{center}
\includegraphics[width=.95\linewidth]{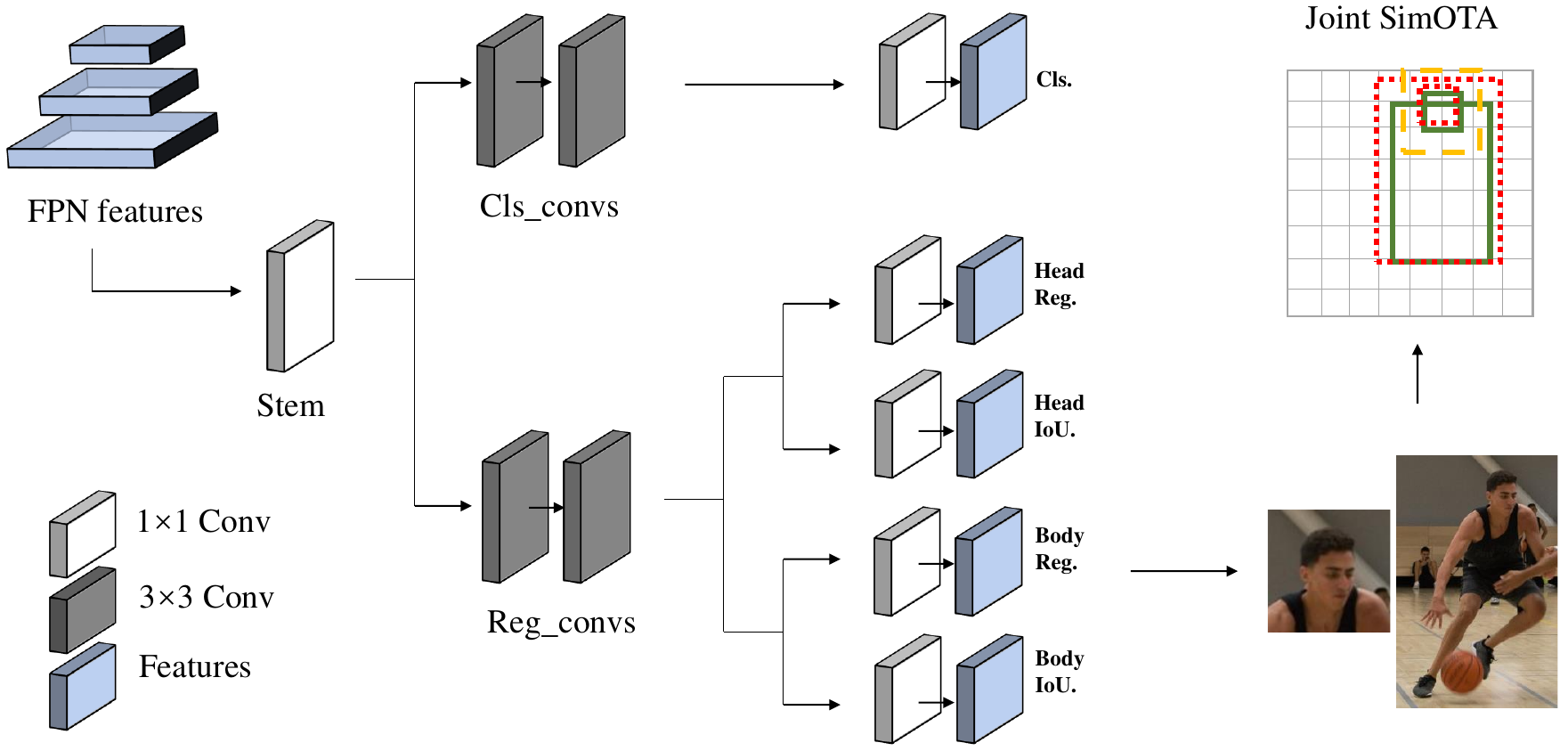}
\end{center}
   \caption{Overview of the proposed model. The image is first processed by a FPN structure to extract features of different scales. Then, a 1 $\times$ 1 stem conv layer followed by two 3 $\times$ 3 branches extract the classification and regression features of the image. For the regression branch, we add two additional prediction heads to predict the head and body simultaneously. The predictions are evaluated by a Joint SimOTA module to filter out the grids that have either bad head prediction or bad body prediction. We calculate losses for the rest of the grids.}
\label{fig:overview}
\end{figure*}

\section{Related Work}
\subsection{Object detection and occlusion handling in pedestrian tracking}
Pedestrian detection has been widely studied for its significance in many real-world applications such as indoor localisation, crowd control and autonomous driving. Since Convolution Neural Networks (CNNs) are introduced to the field, many successful two-stage benchmarks have been proposed \cite{ren2015faster, yang2016exploit}. They first use a Region Proposal Network (RPN) to find out possible regions for pedestrians and then use CNN to predict their location. Another popular approach is the one stage detector represented by YOLO \cite{redmon2016you}. They divide the image into an $S \times S$ grids and directly predict pedestrians from the predefined anchors in those grids. Recent researches suggest that removing the anchors from the detection process help to improve the accuracy and significantly reduce the inference time \cite{tian2019fcos, zhou2019objects}. Despite the development of new detection methods, occlusion has been a big challenge to further improve the performance. People overlapping with each other, hiding behind obstacles and making different poses cause a lot of miss detection and inaccuracy.

To handle a variety of occlusions, a general framework is to integrate body detector with the pedestrian detector. Early research like DeepParts \cite{tian2015deep} handle occlusion
with a part pool and train extensive part detectors to improve the pedestrian recall rate. Zhou and Yuan \cite{zhou2016learning} categorize different types of occlusion patterns and use its own occlusion-specific detectors that are trained simultaneously to obtain the final classifier. Bi-box \cite{zhou2018bi} use two separate detection branches in the model to detect the visible part and the full body at the same time. Chu et al. \cite{chu2020detection} propose a new framework to predict multiple instances instead of one and introduce a new EMD Loss and Set NMS.

When applying occlusion handling techniques to MOT, more features can be explored. SORT \cite{bewley2016simple} leverages the motion factor to handle occlusions by predicting the pedestrian locations in the next frame and match them with the incoming detection. Visual feature is another important factor. DeepSort \cite{wojke2017simple} explores the ReID features as part of the matching metric. FairMOT \cite{zhang2021fairmot} proposes a joint detect-tracking framework that can extract the ReID features while detecting the pedestrians. MTrack \cite{yu2022towards} improves FairMOT by extracting the visual features from multiple body parts instead of the body center. However, these approaches all have limited performance in heavily occluded scenes because the ReID features for pedestrians in such scenes are sometimes inconsistent and unreliable. Therefore, it is considered that the visual features have little impact in today's tracker \cite{zhang2021bytetrack} as the tracker just relies on exploiting the potential of low score detection and matching them with high score detection with a traditional Kalman filter, which after all could still achieves state-of-the-art performance.

\subsection{Head and body detection in dense crowd tracking}

Head detection is difficult due to the small size of head compared to pedestrian body, hence it is firstly used for crowd counting \cite{zhang2015cross, zhang2016single} by estimating the crowd density. The development of Feature Pyramid Network (FPN) \cite{lin2017feature} enables a more accurate detection of objects in different scales. Since head detection scenes are generally more challenging than body detection scenes in terms of crowd density, many researchers have discovered that combining head detection with body detection usually help to improve the detector performance. 

Lu et al. \cite{lu2020semantic} propose a head-body alignment net to jointly detect human head and body. They use two parallel RPN branches to propose the head body RoIs and use an additional Alignment Loss to enforces body boxes to locate compactly around the head region. PedHunter \cite{chi2020pedhunter} train the model to predict a head mask while predicting the pedestrian bodies. It serves as an attention mechanism to assist the feature learning in CNN. It helps to reduce the false positive by learning more distinguishable pedestrian features. However, the method does not eliminate the false negatives as the RPN in the model is trained just like traditional body detector. Chi et al. propose JointDet\cite{chi2020relational} to use pedestrian heads to assist in the body detection which shares a similar idea to this work. However, they use a static body-head ratio to generate body proposals from head proposals to reduce the computational workload in tiling the anchors for body proposals. In our work, we abandon the previous anchor-based design and propose a much simpler and more efficient solution for the joint head-body detection problem.

\section{Methodology}
\subsection{Framework Overview}
The overall framework is shown in Figure \ref{fig:overview}. For the object detection part, we train a head detector adopting the YOLOX\cite{ge2021yolox} structure. We add an additional regression branch to the decoupled head and the classification head remains the same. The detector aims to classify and regress human heads in the scene. It also generates a body prediction along with each head detection. To combine the main detector with the supplementary body detector in the process of multiple object tracking, we first do a bipartite matching to pair our head detection with the external body detection, then the unmatched head detection can contribute to the tracking process.

\begin{figure*}[t]
% \begin{center}
% \includegraphics[width=.95\linewidth]{fig/supptrack_heatmap-cropped.pdf}
   \begin{subfigure}{.33\textwidth}
   \centering
    \includegraphics[width=\linewidth]{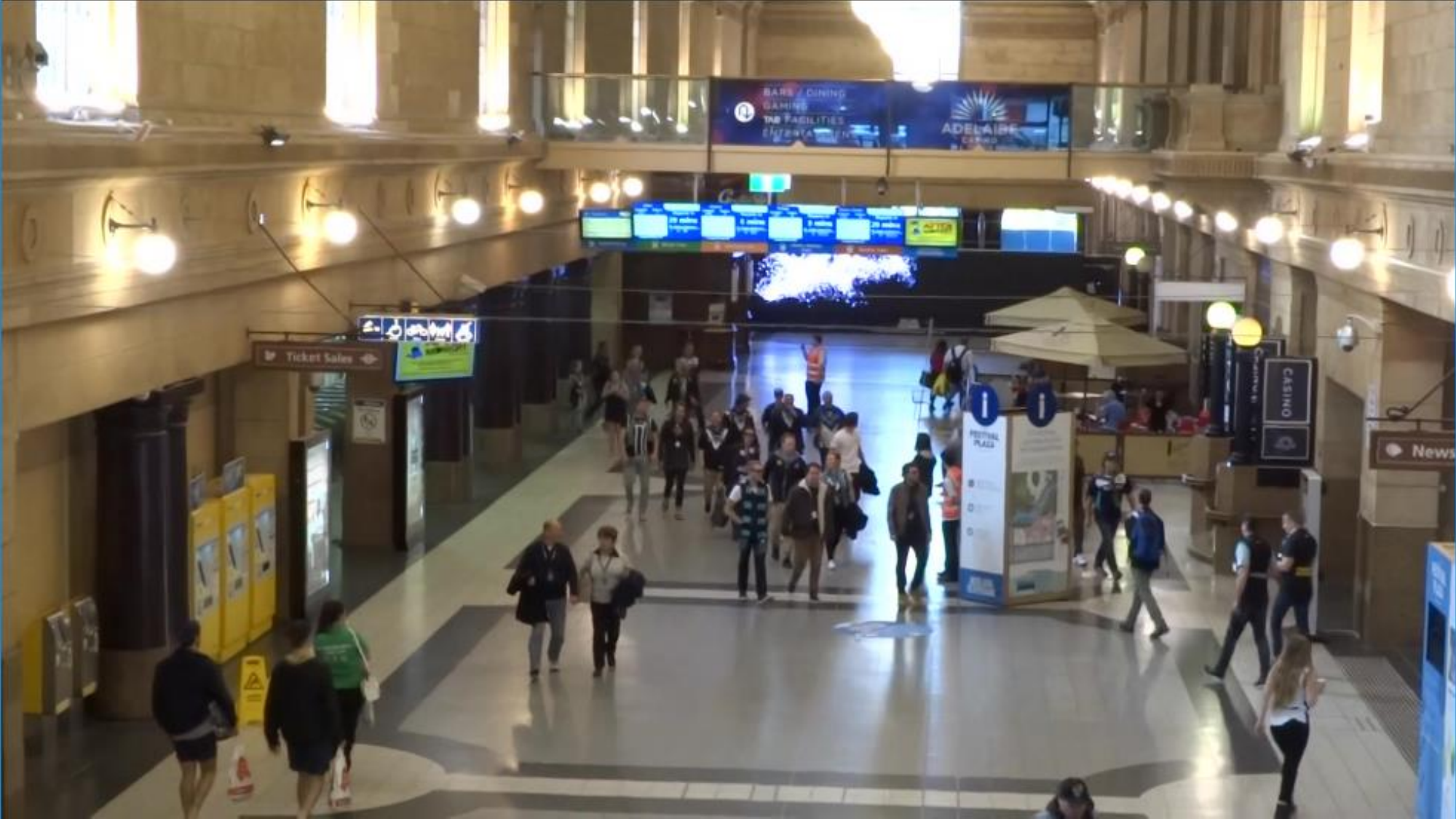}
    \caption{Input image} \label{fig:1a}
  \end{subfigure}%
  \hspace*{\fill}   
  \begin{subfigure}{.33\textwidth}
  \centering
    \includegraphics[width=\linewidth]{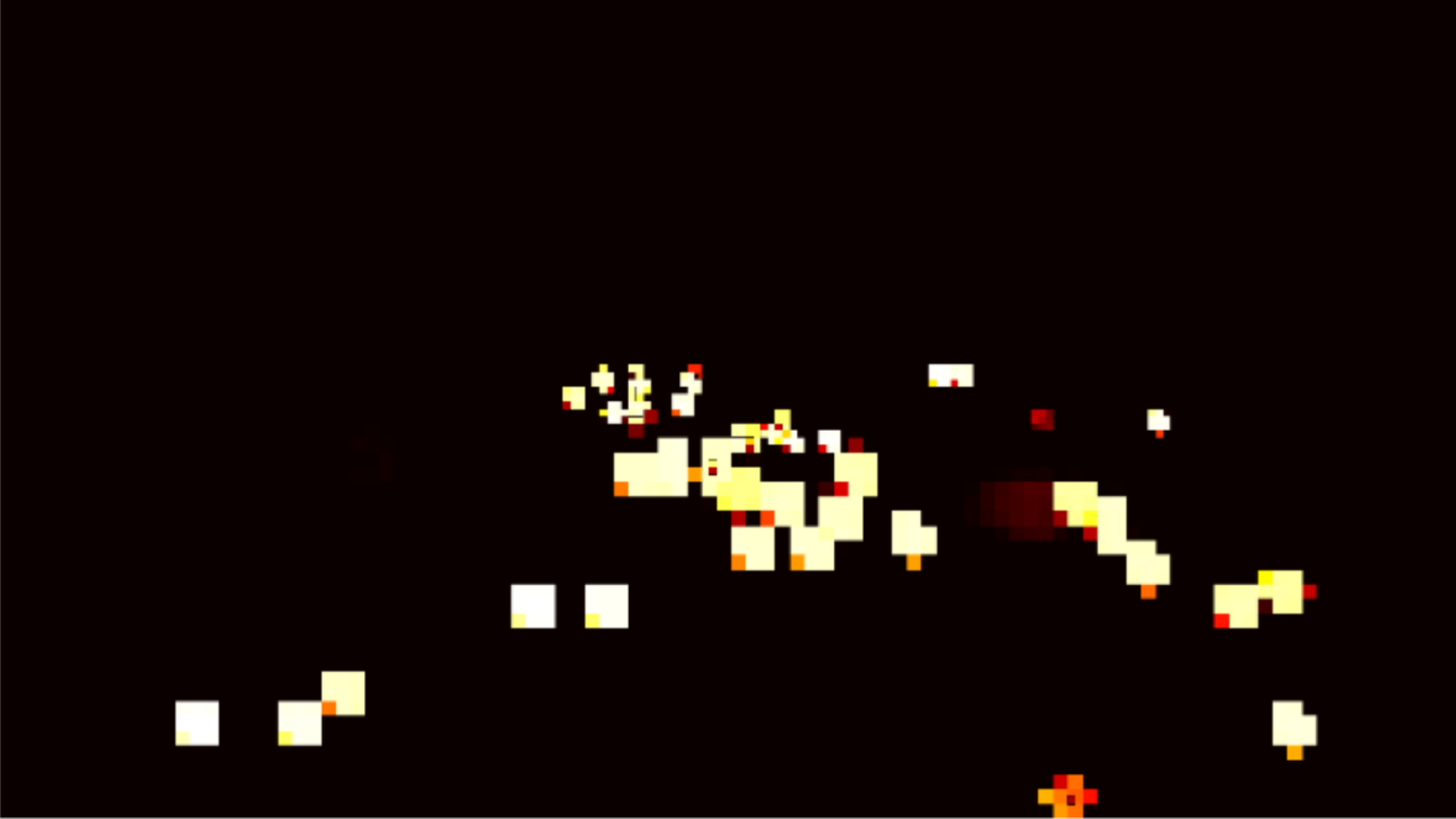}
    \caption{Baseline} \label{fig:1b}
  \end{subfigure}%
  \hspace*{\fill}   
  \begin{subfigure}{.33\textwidth}
  \centering
    \includegraphics[width=\linewidth]{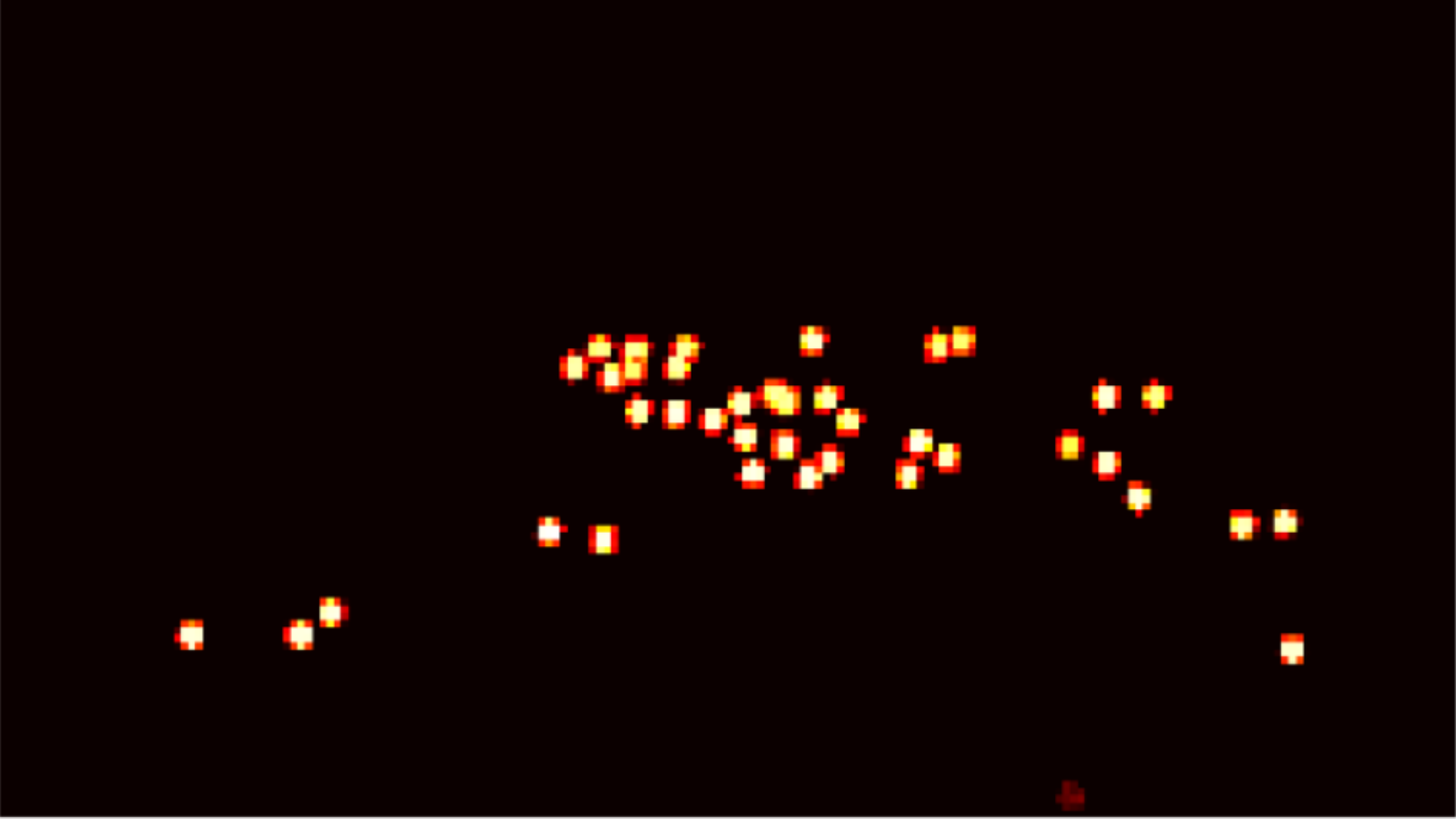}
    \caption{Ours} \label{fig:1c}
  \end{subfigure}
% \end{center}
  \caption{Heatmap comparison between head detection and body detection model under the same model structure. (a) is the input image from MOT20 dataset. (b) is the heatmap from the baseline. (c) is the heatmap from our model. Comparing to the baseline, our heatmap is more distinguishable and can detect more pedestrians in the scene. }
\label{fig:heatmap}
\end{figure*}

A heat map comparison is shown in Fig \ref{fig:heatmap}. For the input image Figure \ref{fig:1a}, we demonstrate two heat maps: one from the baseline body detector and the other is from our joint head-body detector. The two detectors are built with identical backbone network and are trained with the same training dataset \& equal epochs for fair comparison. From the figure, we can see that the head detection model can extract more distinguishable features from the image than the body detection model. The head heat map also contains features that are missed by the baseline body detector (e.g. the man sitting on the right side and the man standing in the middle). Hence, it should perform better on extremely crowded scenes with a proper architecture design.

The advantages of the proposed approach are: 1). To address the head and body proposal relationship problem existed in previous work. Since a person can have various posture, generating the body proposal according to the common standing posture ratio is not always reliable. A bad proposal often leads to a bad prediction. Hence, we solve this issue by training an anchor-free detector and predict the head and body from the same feature maps simultaneously. 2). In crowded scenes, human heads are usually easier to detect. In some heavily occluded scenes, the head of a person can be easily observed while the rest of the body is hidden. However, only detecting heads is not sufficient for MOT problems since sometimes heads are not visible. Thus, we use an additional body detector to serve as a complementary model.

\subsection{Anchor-free head-body detection}
In the previous design of JointDet\cite{chi2020relational}, pedestrian head and body are predicted from two separate RoIs following a specific head-body ratio. This causes two problems: 1). The ratio is statistically obtained based on all human head-body pairs in the CrowdHuman dataset, which are dominantly from a horizontal camera angle, resulting in a standing posture head-body ratio. The body proposal generated from this ratio is not suitable for the top-down camera angle sequences in MOT. 2). The extra hyperparameters introduced by the anchor boxes heavily influences the performance, impacting the overall recall and precision score. 

\begin{figure}[t]
% \begin{center}
% \includegraphics[width=0.9\linewidth]{fig/supptrack_anchor-cropped.pdf}
   \begin{subfigure}{.24\textwidth}
   \centering
    \includegraphics[width=\linewidth]{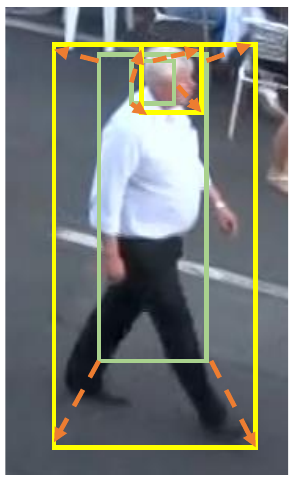}
    \caption{Anchor-based} \label{fig:2a}
  \end{subfigure}%
  \hspace*{\fill}   
  \begin{subfigure}{.237\textwidth}
  \centering
    \includegraphics[width=\linewidth]{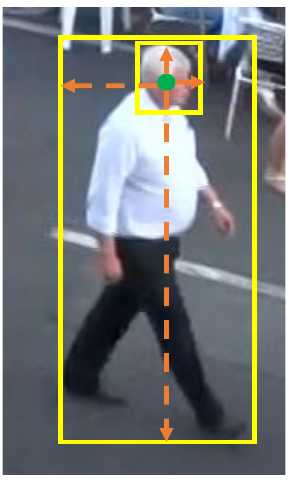}
    \caption{Anchor-free} \label{fig:2b}
  \end{subfigure}%

% \end{center}
   \caption{Illustration of our anchor-free approach vs. traditional static head-body ratio anchor-based approach. a). Anchor-based approach, where head and body boxes are regressed from separate RoIs. b). Our anchor-free approach, where head and body boxes are predicted simultaneously from the same grid box, achieving a faster inference speed and less parameters for training.}
\label{fig:anchor}
\end{figure}

Therefore, we adopt the design of the recent YOLOX \cite{ge2021yolox} and propose an anchor-free style joint head-body detector. As shown in Figure \ref{fig:overview}, for each location, the regression value (top left x and y, width and height) of the head and body are directly predicted by two parallel CNN branches. An illustration is shown in Figure \ref{fig:anchor}.

With the anchor-free design, our model can directly predict the pedestrian body box without using a predefined body anchor box while greatly reducing the requested efforts to heuristically tune the parameters for training. Supported by the training advantage, we also eliminate the performance impact by the body anchors.

\subsection{Joint SimOTA}
We use both the head prediction and body prediction to perform label assignment during training. Specifically, the pair-wise matching degree between a ground truth $g_i$ and prediction $p_j$ is calculated as:
\begin{equation}
    C_{ij} = L_{ij}^{cls} + \lambda_1 L_{ij}^{hreg} + \lambda_2 L_{ij}^{breg}
\end{equation}
where $\lambda_1$ and $\lambda_2$ are the balancing coefficient. $L_{ij}^{cls}$ is the classification loss between the ground truth $g_i$ and prediction $p_j$. $L_{ij}^{hreg}$ and $L_{ij}^{breg}$ are the regression losses for head and body prediction respectively. For a ground truth $g_i$, we select the top $k$ predictions with the least cost within a fixed center region of the head ground truth box as its positive samples. The value $k$ is determined dynamically according to the IoUs of the predicted boxes \cite{ge2021ota}. The grids that contains these positive samples are considered as positive grids and all other grids are consider negative. Only positive grids are saved up for loss computing to reduce the computing resource cost.

The Joint SimOTA makes sure that during the grid sample assignment process, only those with both a good head prediction and a good body prediction are selected as positive samples. It serves as a function similar to the combination of the head-body relationship discriminating module (RDM) and the statistical head-body ratio proposal generation method introduced in \cite{chi2020relational}. Instead of generating the body proposals from heads based on a statistical ratio, all head-body prediction pairs are generated based on the features extracted by the stem network and dynamically evaluated by the joint simOTA process. This helps to simplify the prediction process by reducing the extra hyper-parameters introduced by RDM module and boost the average detection precision.
\subsection{Tracking framework}
We use a simple yet efficient two-step tracking framework inspired by ByteTrack \cite{zhang2021bytetrack}. First, we need to combine the detection results from both detectors to eliminate the duplicated detection and split the detection set into two parts: first class detection and second class detection. We do a bipartite matching between the body prediction from the traditional body detector and our joint head-body detector. This creates three different parts of detection:
\begin{itemize}
  \item Matched detection. This represents the people detected by both the main detector and the supplementary detector.Their confidence scores are set to which one is higher in either detectors. This helps to enhance weak detection scores.
  \item Unmatched body. This represents the people detected by supplementary body detector only, usually happens with pedestrians that are walking out of the screen's top edge since their heads are not visible.
  \item Unmatched head. This represents the pedestrians that are only detected by our joint head-body detector. Usually happens with heavily overlapped crowd whose body are highly occluded but heads are clearly visible.
\end{itemize}

We use Hungarian Algorithm for the bipartite matching step. The matched head and body are all classified as first class detection. For the unmatched head and unmatched body, we further classify them according to their confidence score. Once we have the two sets of detection, we can perform tracking. The detailed tracking algorithm is shown in Algorithm \ref{algo:tracking}.

% Then, we collect all the matched detection and the unmatched body to perform the first round of tracking. We choose the algorithm from \cite{zhang2021bytetrack} as it yields the best result and runs fast. We do the second round of tracking with the unmatched heads only with the same tracking algorithm. Tracking results from these two rounds are summed up to obtain the final tracking results. Some post processing techniques are also used after the tracking is done such as tracklet interpolation and duplicated tracklet removal to further boost performance. The detailed tracking algorithm can be found in algorithm \ref{algo:tracking}.

% More comparison between different tracking approaches are discussed in Section ?.
\begin{algorithm}[htb]

\caption{Tracking algorithm with both detectors}\label{tracking}

\begin{algorithmic}[1]
 \renewcommand{\algorithmicrequire}{\textbf{Input:}}
 \renewcommand{\algorithmicensure}{\textbf{Output:}}
 \REQUIRE head detection $h\_Det$, body detection $b\_Det$
 \ENSURE  tracklets $\tau$
 \\ \textit{Initialisation} : $\tau \leftarrow \emptyset$
%   \STATE first statement
%  \\ \textit{LOOP Process}
  \FOR {frame $i$ in video length}
  \STATE match $h\_Det_{i}$ with $b\_Det_{i}$ 
  \STATE $\mathcal{D}_{high} \leftarrow$ matched heads and bodies, high score heads, high score bodies
  \STATE $\mathcal{D}_{low} \leftarrow$ low score heads, low score bodies
  \\ \textit{------Update all tracklets with Kalman filter------}
  \FOR {$t$ in $\tau$}
  \STATE $t \leftarrow KalmanFilter(t)$
  \ENDFOR
  \\ \textit{---------First class detection association---------}
  \STATE match $\tau$ with $\mathcal{D}_{high}$
  \STATE $\tau_{remain} \leftarrow$ unmatched tracklets
  \STATE $\mathcal{D}_{remain} \leftarrow$ unmatched high score detection
  \\ \textit{---------Second class detection association---------}
  \STATE match $\tau_{remain}$ with $\mathcal{D}_{low}$
  \\ \textit{---------Update and initialize tracklets---------}
  \STATE $\tau \leftarrow$ matched detection
  \STATE $\tau \leftarrow \tau + \mathcal{D}_{remain}$
%   \IF {($i \ne 0$)}
%   \STATE statement..
%   \ENDIF
  \ENDFOR
 \RETURN $\tau$ 
 \end{algorithmic}
 
\label{algo:tracking}
\end{algorithm}

\subsection{Loss Function}
The loss function of the purposed method consists of four parts and is defined as follows:
\begin{equation}
    \mathcal{L} = \mathcal{L}_{cls} + \mathcal{L}_{obj} + \alpha_1\mathcal{L}_{head} + \alpha_2\mathcal{L}_{body}
\end{equation}
where $\mathcal{L}_{cls}$ and $\mathcal{L}_{obj}$ is the binary cross entropy loss for head classification and confidence score prediction. $\mathcal{L}_{head}$ and $\mathcal{L}_{body}$ are the regression loss for the predicted head and body bounding boxes. $\alpha_1$ and $\alpha_2$ are the balancing coefficients and we set them to 5 during our training and experiments. We use an additional L1 loss for the head and body detections for the last 10 epochs of training.

\begin{figure*}[t]
\begin{center}
\includegraphics[width=\linewidth]{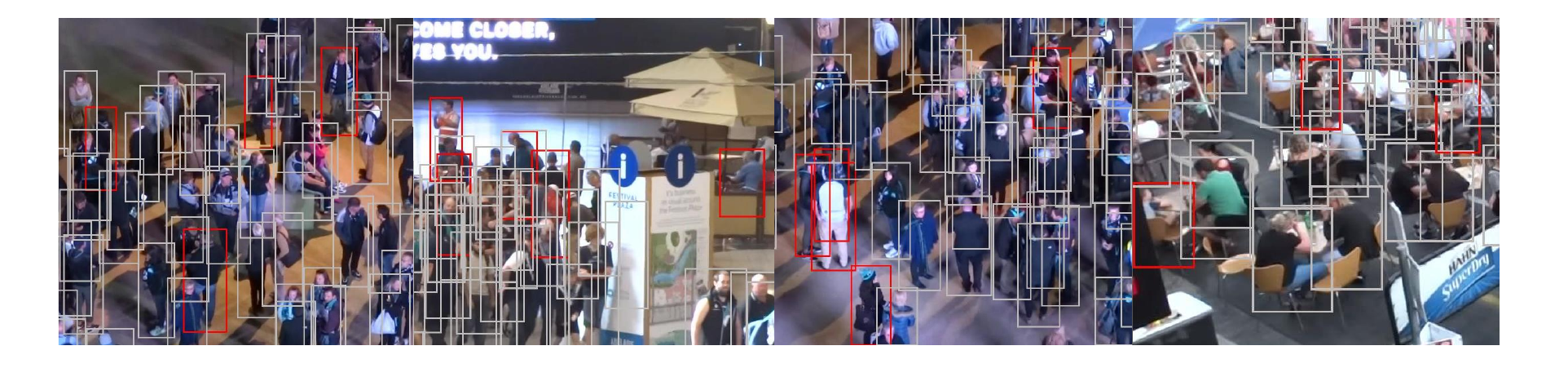}
\end{center}
   \caption{Qualitative results on MOT20 dataset. Pedestrians detected by the baseline are shown in grey boxes. Pedestrians detected by our method which are not detected by the baseline are shown in red boxes. Best viewed in colour.}
\label{fig:qualitative}
\end{figure*}

\subsection{Training Details}
\subsubsection{Dataset construction and data augmentation} Since there is few dataset containing head and body labels at the same time, we use existing datasets to build our own training data. We combine the ground truth for MOT20 and HT21 since they have two identical training sequences. We also use the Crowdhuman dataset during training since they explicitly label the pedestrian heads and bodies. Following the work from \cite{ge2021yolox}, we use Mosaic and MixUp strategies to boost the detecting performance. They are commonly used in YOLOv4 \cite{bochkovskiy2020yolov4}, YOLOv5 \cite{glenn_jocher_2020_4154370}, and other detectors.
\subsubsection{Training parameters setting}
We use a YOLOX-x model pretrained on COCO dataset to initialize the training. The model was trained on two RTX3090 graphic cards with a batch size of 8 and epoch number of 80. Following the design of ByteTrack\cite{zhang2021bytetrack}, the optimizer is SGD with a momentum ratio of 0.9. The learning rate is set to 0.0001 with 1 epoch warmup and use the cosine annealing strategy introduced in \cite{loshchilov2016sgdr}. The input of the training images are set to $896\times1600$. The entire training takes about 42 hours. For post processing, we use a NMS of threshold of 0.45 for the head prediction and 0.7 for the body prediction to eliminate overlapped detection.

\begin{figure}[t]
\begin{center}
\begin{subfigure}{.24\textwidth}
   \centering
    \includegraphics[width=\linewidth]{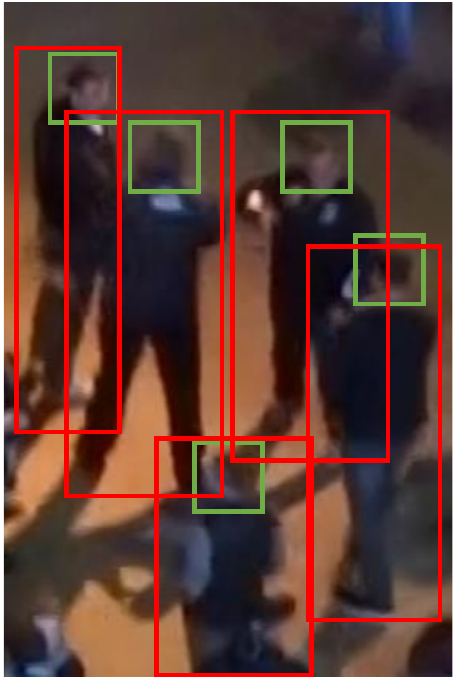}
    \caption{} \label{fig:2a}
  \end{subfigure}%
  \hspace*{\fill}   
  \begin{subfigure}{.24\textwidth}
  \centering
    \includegraphics[width=\linewidth]{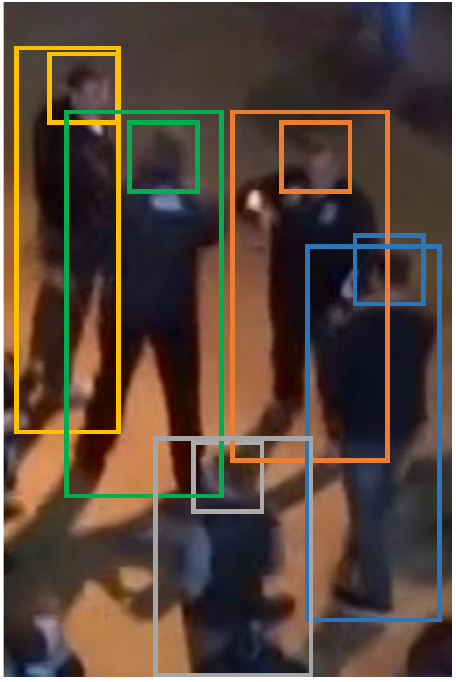}
    \caption{} \label{fig:2b}
  \end{subfigure}%
   
\end{center}
   \caption{Demo of creation of our mixed training dataset. We use bipartite matching to assign head labels to body labels. a): Red boxes represents the body labels from MOT20 dataset, green boxes represents the head labels from HT21 dataset; b): after bipartite matching, each head label is assigned a body label similar to Crowdhuman dataset. Same colour represents the label box belong to the same person}
\label{fig:dataset}
\end{figure}

\section{Experiments}
Since our approach focus on extremely crowded scenes, the experiments are mainly conducted on MOT20 dataset. Experiments are also conducted on Crowdhuman and HT21 to demonstrate the robustness of the proposed approach.

\subsection{MOT Challenge}
MOTChallenge is a human tracking benchmark that provides carefully annotated datasets and clear metrics to evaluate the performance of tracking algorithms and pedestrian detectors. Earlier benchmark MOT17 provides over 29$k$ annotations in a total of 14 sequences. MOT20 is a more recent benchmark that consist of 8 different sequences depicting very crowded challenging scenes.
\subsubsection{MOT20}
For the MOT20 challenge, we conduct two parts of experiments. The first one is the detection performance and the second one is the tracking performance. Similar to the "Pedestrian Detection in Complex and Crowded Events" in HiEve, we evaluate the detection performance on MOT20 dataset. Since the ground truth label for the MOT20 test data is not available, we conduct the experiment on the training set and show qualitative for the test set. 
% For a fair comparison, we retrain the models, using only the training set from CrowdHuman and exclude MOT20 from the training data. 
The training dataset for our experiment is the Crowdhuman dataset and the mixed dataset we created for our model. We choose two sequences from the HT21 dataset and the MOT20 dataset and combine their ground truth label. We add offset to the label to match the image size differences. Figure \ref{fig:dataset} is a dataset sample.

For the detection performance metric, we choose the log-average miss rate over 9 points ranging from $10^{-2}$ to $10^{0}$ FPPI (the MR$^{-2}$) to evaluate the performance of our detector. The performance comparison is shown in Table \ref{tab:mot20_det}. We use current top-performance detector used in ByteTrack \cite{zhang2021bytetrack} as our baseline. As shown in the table, our method achieves 92.99\% AP which is 7.32\% higher than the baseline. For the log average missing rate, we also achieves 7\% MR$^{-2}$ which is 7\% better than the base line. The result has shown that our method reaches the supreme pedestrian detection performance by letting the model focus on the heads instead of the bodies. 

\begin{table}[tbp]
\caption{AP and MR$^{-2}$ performance of different detection methods on MOT20}

\centering
\resizebox{0.8\linewidth}{!}{
% \begin{tabular}{ |c|c|c| } 
%  \hline
%  Method & AP$\uparrow$ & $MR^{-2}\downarrow$ \\ 
%  \hline
%  ByteTrack & 85.67 & 14.0 \\ 
%  Ours & 92.99 & 7.0 \\ 
%  \hline
%  \label{tab:mot20_det}
% \end{tabular}

\begin{tabular}{ |c|c|c| } 
 \hline
 Method & AP$\uparrow$ & MR$^{-2}\downarrow$ \\ 
  \hline
 Baseline & 85.67 & 14.0 \\ 
 Ours w/o Joint SimOTA & 91.16 & 9.0\\
 Ours & 92.99 & 7.0 \\ 
 \hline
\end{tabular}
}
\label{tab:mot20_det}
\end{table}

For the tracking performance, the results are uploaded to the official website for evaluation. The main evaluation metrics are Multiple Object Tracking Accuracy (MOTA), IDF1 and Higher Order Tracking Accuracy (HOTA). The result is compared with FairMOT \cite{zhang2021fairmot}, TransCenter \cite{xu2021transcenter}, TransTrack \cite{sun2020transtrack}, CSTrack \cite{liang2022rethinking}, SOTMOT \cite{zheng2021improving}, MAA \cite{stadler2022modelling}, ReMOT \cite{yang2021remot} and ByteTrack \cite{zhang2021bytetrack}. Despite many detections are ignored in the process of evaluation \cite{dendorfer2020mot20}, we still achieve state-of-the-art performance of 78.2\% MOTA and 75.5\% IDF1. 

\begin{table}[tbp]
\caption{Performance comparison on the test set with state-of-the-art on MOT20}

\centering
\resizebox{\linewidth}{!}{\begin{tabular}{lccccccc}
\toprule
. & MOTA$\uparrow$ & IDF1$\uparrow$ & HOTA$\uparrow$  & FP$\downarrow$ & FN$\downarrow$ & IDs$\downarrow$ \\
\midrule
FairMOT & 61.8 &  67.3 & 54.6 &  103440 &  88901 & 5243 \\
TransCenter & 61.9 &  50.4 & 43.5 & 45895 &  146347 & 4653 \\
TransTrack & 65.0 &  59.4 & 48.5 & 27197 & 150197 & 3608 \\
CSTrack & 66.6 & 68.6 & 54.0 & 25404 & 144358 & 3196 \\
SOTMOT & 68.6 &  71.4 & 57.4 & 57064 & 101154 & 4209 \\
MAA & 73.9 & 71.2 & 57.3 & \textbf{24942} & 108744 & 1331 \\
ReMOT & 77.4 & 73.1 & 61.2 & 28351 & 86659 & 2121  \\
ByteTrack & 77.8 & 75.2	 & 61.3 & 26249 & 87594 & \textbf{1223}   \\
Ours & \textbf{78.2} & \textbf{75.5} & \textbf{61.9} & 30187 & \textbf{81119} & 1325 \\
\bottomrule
\end{tabular}}
\end{table}

\subsection{Qualitative Result on MOT20}

We demonstrate the effectiveness of our model in this section. The results are shown in Figure \ref{fig:qualitative}. We have randomly selected 3 tracking sequences from the MOT20 test set and perform detection with both the baseline approach and our proposed approach. It is clear that from Figure \ref{fig:qualitative}, our approach has a stronger ability to detect pedestrians in crowded scenes. In Figure \ref{fig:qualitative}, the red boxes are the pedestrians only detected by our method not from the baseline. It usually happens to people that are heavily overlapped with others or only showing part of their body due to different postures. Since our model focuses on detecting the heads, it can extract the corresponding head features from the image and accurately predict the body boxes.

\begin{table}[tbp]
\caption{Performance comparison on the test set with state-of-the-art on HT21}

\centering
\resizebox{\linewidth}{!}{\begin{tabular}{lccccccc}
\toprule
. & MOTA$\uparrow$ & IDF1$\uparrow$ & IDEucl$\uparrow$  & HOTA$\downarrow$ & FP$\downarrow$ & FN$\downarrow$ \\
\midrule
HeadHunter & 57.8 &  53.9 & 54.2 &  36.8 &  51840 & 299459 \\
FairMOT & 60.8 &  62.8 & 69.9 &  43.0 &  118109 & 198896 \\
FM\_OCSORT & 67.9 &  62.9 & 62.1 & 44.1 &  102050 & 164090 \\
THT & 70.7 &  68.4 & 63.5 & 47.3 & 33545 & 211162 \\
pptracking & 72.6 & 61.8 & 59.7 & 44.6 & 71235 & 154139 \\
Ours & \textbf{77.9} & \textbf{70.8} & \textbf{68.4} & \textbf{50.2} & \textbf{42,205} & \textbf{140867} \\
\bottomrule
\end{tabular}}
\label{tab:ht21}
\end{table}

\begin{table}[tbp]
\caption{head and body AP performance of different detection methods on Crowdhuman}

\centering
\resizebox{0.6\linewidth}{!}{

\begin{tabular}{ |c|c|c| } 
 \hline
 Method & head & body \\ 
  \hline
 Baseline-Head & 54.2 & - \\ 
 Baseline-Body & - & 57.8 \\ 
 Ours & 55.0 & 57.9 \\ 
 \hline
\end{tabular}
}
\label{tab:crowdhuman}
\end{table}

\subsection{Ablation study on Joint SimOTA}

% 47.9AP 
% 48.3AP 
Our Joint SimOTA module helps the model to learn head-body relationship dynamically which can improve the detection performance. We conduct ablation study by training our proposed model with the same settings except the Joint SimOTA module. According to Table \ref{tab:mot20_det}, adding the Joint SimOTA module can boost the AP by 1.83\% and reduce the log average missing rate by 2\%. The performance improvement indicates that body regression contributing to the training process helps the model to predict more accurate body bounding boxes. Otherwise, the model learns features for the head and body separately leading to more false positives.

\subsection{Crowdhuman}
% bring out head detection as well?
Crowdhuman is a public pedestrian detection dataset that contains various crowded scenes. It provides over 470$k$ pedestrian labels in a total of 15000 training set and 4370 validation set. For each pedestrian, the label contains the head, visible body part and full body annotation. To demonstrate the robustness of our method, we conduct experiments on the validation set with both the head and body detection. We retrain the YOLOX model for the two tasks separately as our baseline. According to Table \ref{tab:crowdhuman}, our approach improves the head and body AP by 0.8\% and 0.1\% respectively. The result shows that combining the head and body detection helps to boost the performance of both tasks.

\begin{figure}[t]
% \begin{center}
% \includegraphics[width=0.9\linewidth]{fig/supptrack_anchor-cropped.pdf}
   \begin{subfigure}{.155\textwidth}
   \centering
    \includegraphics[width=\linewidth]{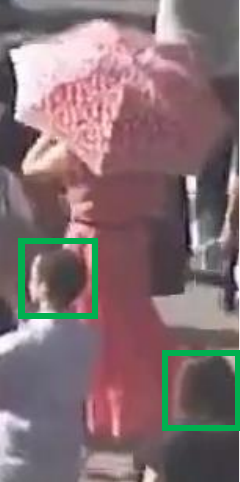}
    \caption{} \label{fig:2a}
  \end{subfigure}%
  \hspace*{\fill}   
  \begin{subfigure}{.155\textwidth}
  \centering
    \includegraphics[width=\linewidth]{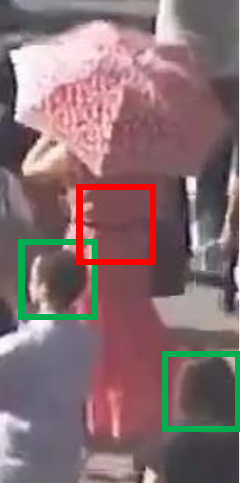}
    \caption{} \label{fig:2b}
  \end{subfigure}%
  \begin{subfigure}{.157\textwidth}
  \centering
    \includegraphics[width=\linewidth]{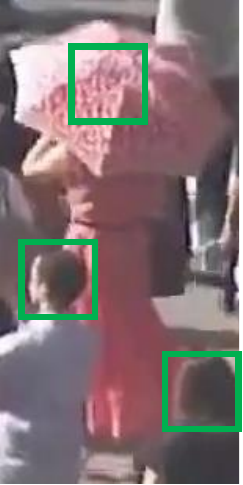}
    \caption{} \label{fig:2b}
  \end{subfigure}%

% \end{center}
   \caption{Head detection comparison between our model and baseline model. The person dressed in pink covered their head with an umbrella, causing the head features invisible to the CNN. a). One of the baseline model only detects two heads in the scene; b). Another baseline model predicts wrong location of the head, represented with a red box; c). Our model can correctly predict all three head location of the person.}
\label{fig:head}
\end{figure}

\subsection{Head detection and tracking}
Since our model detects head and body simultaneously, the model can also be adopted in head tracking datasets. Because of the Joint SimOTA module we added in the training phase, the body part of a pedestrian can contribute to the head prediction as well. As shown in Figure \ref{fig:head}, the pedestrian's head is covered by an umbrella. The baseline models search for the head features in the figure and end up with missing or wrong detection. Our proposed model can correctly predict the head location of the pedestrian. Compared to HeadHunter \cite{sundararaman2021tracking}, FairMOT \cite{zhang2021fairmot}, FM\_OCSORT \cite{cao2022observation}, THT and pptracking \cite{ppdet2019}, our joint head-body detection model achieves state-of-the-art results on the HT21 head tracking challenge without needing for additional training data as shown in Table \ref{tab:ht21}.

\section{Conclusion}
In this paper, we propose an anchor-free style joint head-body detection model to detect pedestrians' head and body simultaneously. By focusing on the heads, the model can detect pedestrians more effectively than the baseline body detectors due to high occlusion. With our proposed model, we can significantly reduce the detected false negative in extremely crowded scenes. We have conducted extensive experiments on MOT20, HT21 and Crowdhuman datasets. Our approach achieves state-of-the-art detection and tracking performance and is robust to various scenarios.

\bibliography{aaai22}

\begin{thebibliography}{38}
\providecommand{\natexlab}[1]{#1}

\bibitem[{Authors(2019)}]{ppdet2019}
Authors, P. 2019.
\newblock PaddleDetection, Object detection and instance segmentation toolkit
  based on PaddlePaddle.
\newblock \url{https://github.com/PaddlePaddle/PaddleDetection}.

\bibitem[{Bewley et~al.(2016)Bewley, Ge, Ott, Ramos, and
  Upcroft}]{bewley2016simple}
Bewley, A.; Ge, Z.; Ott, L.; Ramos, F.; and Upcroft, B. 2016.
\newblock Simple online and realtime tracking.
\newblock In \emph{2016 IEEE international conference on image processing
  (ICIP)}, 3464--3468. IEEE.

\bibitem[{Bochkovskiy, Wang, and Liao(2020)}]{bochkovskiy2020yolov4}
Bochkovskiy, A.; Wang, C.-Y.; and Liao, H.-Y.~M. 2020.
\newblock Yolov4: Optimal speed and accuracy of object detection.
\newblock \emph{arXiv preprint arXiv:2004.10934}.

\bibitem[{Cao et~al.(2022)Cao, Weng, Khirodkar, Pang, and
  Kitani}]{cao2022observation}
Cao, J.; Weng, X.; Khirodkar, R.; Pang, J.; and Kitani, K. 2022.
\newblock Observation-Centric SORT: Rethinking SORT for Robust Multi-Object
  Tracking.
\newblock \emph{arXiv preprint arXiv:2203.14360}.

\bibitem[{Chi et~al.(2020{\natexlab{a}})Chi, Zhang, Xing, Lei, Li, and
  Zou}]{chi2020pedhunter}
Chi, C.; Zhang, S.; Xing, J.; Lei, Z.; Li, S.~Z.; and Zou, X.
  2020{\natexlab{a}}.
\newblock Pedhunter: Occlusion robust pedestrian detector in crowded scenes.
\newblock In \emph{Proceedings of the AAAI Conference on Artificial
  Intelligence}, volume~34, 10639--10646.

\bibitem[{Chi et~al.(2020{\natexlab{b}})Chi, Zhang, Xing, Lei, Li, and
  Zou}]{chi2020relational}
Chi, C.; Zhang, S.; Xing, J.; Lei, Z.; Li, S.~Z.; and Zou, X.
  2020{\natexlab{b}}.
\newblock Relational learning for joint head and human detection.
\newblock In \emph{Proceedings of the AAAI Conference on Artificial
  Intelligence}, volume~34, 10647--10654.

\bibitem[{Chu et~al.(2020)Chu, Zheng, Zhang, and Sun}]{chu2020detection}
Chu, X.; Zheng, A.; Zhang, X.; and Sun, J. 2020.
\newblock Detection in crowded scenes: One proposal, multiple predictions.
\newblock In \emph{Proceedings of the IEEE/CVF Conference on Computer Vision
  and Pattern Recognition}, 12214--12223.

\bibitem[{Dendorfer et~al.(2020)Dendorfer, Rezatofighi, Milan, Shi, Cremers,
  Reid, Roth, Schindler, and Leal-Taix{\'e}}]{dendorfer2020mot20}
Dendorfer, P.; Rezatofighi, H.; Milan, A.; Shi, J.; Cremers, D.; Reid, I.;
  Roth, S.; Schindler, K.; and Leal-Taix{\'e}, L. 2020.
\newblock Mot20: A benchmark for multi object tracking in crowded scenes.
\newblock \emph{arXiv preprint arXiv:2003.09003}.

\bibitem[{Ge et~al.(2021{\natexlab{a}})Ge, Liu, Li, Yoshie, and
  Sun}]{ge2021ota}
Ge, Z.; Liu, S.; Li, Z.; Yoshie, O.; and Sun, J. 2021{\natexlab{a}}.
\newblock Ota: Optimal transport assignment for object detection.
\newblock In \emph{Proceedings of the IEEE/CVF Conference on Computer Vision
  and Pattern Recognition}, 303--312.

\bibitem[{Ge et~al.(2021{\natexlab{b}})Ge, Liu, Wang, Li, and
  Sun}]{ge2021yolox}
Ge, Z.; Liu, S.; Wang, F.; Li, Z.; and Sun, J. 2021{\natexlab{b}}.
\newblock Yolox: Exceeding yolo series in 2021.
\newblock \emph{arXiv preprint arXiv:2107.08430}.

\bibitem[{He et~al.(2017)He, Gkioxari, Doll{\'a}r, and Girshick}]{he2017mask}
He, K.; Gkioxari, G.; Doll{\'a}r, P.; and Girshick, R. 2017.
\newblock Mask r-cnn.
\newblock In \emph{Proceedings of the IEEE international conference on computer
  vision}, 2961--2969.

\bibitem[{Jocher(2020)}]{glenn_jocher_2020_4154370}
Jocher, G. 2020.
\newblock {ultralytics/yolov5: v3.1 - Bug Fixes and Performance Improvements}.
\newblock \url{https://github.com/ultralytics/yolov5}.

\bibitem[{Liang et~al.(2022)Liang, Zhang, Zhou, Li, Zhu, and
  Hu}]{liang2022rethinking}
Liang, C.; Zhang, Z.; Zhou, X.; Li, B.; Zhu, S.; and Hu, W. 2022.
\newblock Rethinking the competition between detection and ReID in multiobject
  tracking.
\newblock \emph{IEEE Transactions on Image Processing}, 31: 3182--3196.

\bibitem[{Lin et~al.(2017)Lin, Doll{\'a}r, Girshick, He, Hariharan, and
  Belongie}]{lin2017feature}
Lin, T.-Y.; Doll{\'a}r, P.; Girshick, R.; He, K.; Hariharan, B.; and Belongie,
  S. 2017.
\newblock Feature pyramid networks for object detection.
\newblock In \emph{Proceedings of the IEEE conference on computer vision and
  pattern recognition}, 2117--2125.

\bibitem[{Liu et~al.(2016)Liu, Anguelov, Erhan, Szegedy, Reed, Fu, and
  Berg}]{liu2016ssd}
Liu, W.; Anguelov, D.; Erhan, D.; Szegedy, C.; Reed, S.; Fu, C.-Y.; and Berg,
  A.~C. 2016.
\newblock Ssd: Single shot multibox detector.
\newblock In \emph{European conference on computer vision}, 21--37. Springer.

\bibitem[{Loshchilov and Hutter(2016)}]{loshchilov2016sgdr}
Loshchilov, I.; and Hutter, F. 2016.
\newblock Sgdr: Stochastic gradient descent with warm restarts.
\newblock \emph{arXiv preprint arXiv:1608.03983}.

\bibitem[{Lu, Ma, and Wang(2020)}]{lu2020semantic}
Lu, R.; Ma, H.; and Wang, Y. 2020.
\newblock Semantic head enhanced pedestrian detection in a crowd.
\newblock \emph{Neurocomputing}, 400: 343--351.

\bibitem[{Redmon et~al.(2016)Redmon, Divvala, Girshick, and
  Farhadi}]{redmon2016you}
Redmon, J.; Divvala, S.; Girshick, R.; and Farhadi, A. 2016.
\newblock You only look once: Unified, real-time object detection.
\newblock In \emph{Proceedings of the IEEE conference on computer vision and
  pattern recognition}, 779--788.

\bibitem[{Ren et~al.(2015)Ren, He, Girshick, and Sun}]{ren2015faster}
Ren, S.; He, K.; Girshick, R.; and Sun, J. 2015.
\newblock Faster r-cnn: Towards real-time object detection with region proposal
  networks.
\newblock \emph{Advances in neural information processing systems}, 28.

\bibitem[{Stadler and Beyerer(2022)}]{stadler2022modelling}
Stadler, D.; and Beyerer, J. 2022.
\newblock Modelling ambiguous assignments for multi-person tracking in crowds.
\newblock In \emph{Proceedings of the IEEE/CVF Winter Conference on
  Applications of Computer Vision}, 133--142.

\bibitem[{Sun et~al.(2020)Sun, Cao, Jiang, Zhang, Xie, Yuan, Wang, and
  Luo}]{sun2020transtrack}
Sun, P.; Cao, J.; Jiang, Y.; Zhang, R.; Xie, E.; Yuan, Z.; Wang, C.; and Luo,
  P. 2020.
\newblock Transtrack: Multiple object tracking with transformer.
\newblock \emph{arXiv preprint arXiv:2012.15460}.

\bibitem[{Sundararaman et~al.(2021)Sundararaman, De~Almeida~Braga, Marchand,
  and Pettre}]{sundararaman2021tracking}
Sundararaman, R.; De~Almeida~Braga, C.; Marchand, E.; and Pettre, J. 2021.
\newblock Tracking pedestrian heads in dense crowd.
\newblock In \emph{Proceedings of the IEEE/CVF Conference on Computer Vision
  and Pattern Recognition}, 3865--3875.

\bibitem[{Tian et~al.(2015)Tian, Luo, Wang, and Tang}]{tian2015deep}
Tian, Y.; Luo, P.; Wang, X.; and Tang, X. 2015.
\newblock Deep learning strong parts for pedestrian detection.
\newblock In \emph{Proceedings of the IEEE international conference on computer
  vision}, 1904--1912.

\bibitem[{Tian et~al.(2019)Tian, Shen, Chen, and He}]{tian2019fcos}
Tian, Z.; Shen, C.; Chen, H.; and He, T. 2019.
\newblock Fcos: Fully convolutional one-stage object detection.
\newblock In \emph{Proceedings of the IEEE/CVF international conference on
  computer vision}, 9627--9636.

\bibitem[{Wojke, Bewley, and Paulus(2017)}]{wojke2017simple}
Wojke, N.; Bewley, A.; and Paulus, D. 2017.
\newblock Simple online and realtime tracking with a deep association metric.
\newblock In \emph{2017 IEEE international conference on image processing
  (ICIP)}, 3645--3649. IEEE.

\bibitem[{Xu et~al.(2021)Xu, Ban, Delorme, Gan, Rus, and
  Alameda-Pineda}]{xu2021transcenter}
Xu, Y.; Ban, Y.; Delorme, G.; Gan, C.; Rus, D.; and Alameda-Pineda, X. 2021.
\newblock Transcenter: Transformers with dense queries for multiple-object
  tracking.
\newblock \emph{arXiv preprint arXiv:2103.15145}.

\bibitem[{Yang et~al.(2021)Yang, Chang, Sakti, Wu, and
  Nakamura}]{yang2021remot}
Yang, F.; Chang, X.; Sakti, S.; Wu, Y.; and Nakamura, S. 2021.
\newblock Remot: A model-agnostic refinement for multiple object tracking.
\newblock \emph{Image and Vision Computing}, 106: 104091.

\bibitem[{Yang, Choi, and Lin(2016)}]{yang2016exploit}
Yang, F.; Choi, W.; and Lin, Y. 2016.
\newblock Exploit all the layers: Fast and accurate cnn object detector with
  scale dependent pooling and cascaded rejection classifiers.
\newblock In \emph{Proceedings of the IEEE conference on computer vision and
  pattern recognition}, 2129--2137.

\bibitem[{Yu, Li, and Han(2022)}]{yu2022towards}
Yu, E.; Li, Z.; and Han, S. 2022.
\newblock Towards Discriminative Representation: Multi-view Trajectory
  Contrastive Learning for Online Multi-object Tracking.
\newblock In \emph{Proceedings of the IEEE/CVF Conference on Computer Vision
  and Pattern Recognition}, 8834--8843.

\bibitem[{Zhang et~al.(2015)Zhang, Li, Wang, and Yang}]{zhang2015cross}
Zhang, C.; Li, H.; Wang, X.; and Yang, X. 2015.
\newblock Cross-scene crowd counting via deep convolutional neural networks.
\newblock In \emph{Proceedings of the IEEE conference on computer vision and
  pattern recognition}, 833--841.

\bibitem[{Zhang et~al.(2020)Zhang, Chi, Yao, Lei, and Li}]{zhang2020bridging}
Zhang, S.; Chi, C.; Yao, Y.; Lei, Z.; and Li, S.~Z. 2020.
\newblock Bridging the gap between anchor-based and anchor-free detection via
  adaptive training sample selection.
\newblock In \emph{Proceedings of the IEEE/CVF conference on computer vision
  and pattern recognition}, 9759--9768.

\bibitem[{Zhang et~al.(2021{\natexlab{a}})Zhang, Sun, Jiang, Yu, Yuan, Luo,
  Liu, and Wang}]{zhang2021bytetrack}
Zhang, Y.; Sun, P.; Jiang, Y.; Yu, D.; Yuan, Z.; Luo, P.; Liu, W.; and Wang, X.
  2021{\natexlab{a}}.
\newblock Bytetrack: Multi-object tracking by associating every detection box.
\newblock \emph{arXiv preprint arXiv:2110.06864}.

\bibitem[{Zhang et~al.(2021{\natexlab{b}})Zhang, Wang, Wang, Zeng, and
  Liu}]{zhang2021fairmot}
Zhang, Y.; Wang, C.; Wang, X.; Zeng, W.; and Liu, W. 2021{\natexlab{b}}.
\newblock Fairmot: On the fairness of detection and re-identification in
  multiple object tracking.
\newblock \emph{International Journal of Computer Vision}, 129(11): 3069--3087.

\bibitem[{Zhang et~al.(2016)Zhang, Zhou, Chen, Gao, and Ma}]{zhang2016single}
Zhang, Y.; Zhou, D.; Chen, S.; Gao, S.; and Ma, Y. 2016.
\newblock Single-image crowd counting via multi-column convolutional neural
  network.
\newblock In \emph{Proceedings of the IEEE conference on computer vision and
  pattern recognition}, 589--597.

\bibitem[{Zheng et~al.(2021)Zheng, Tang, Chen, Zhu, Wang, and
  Lu}]{zheng2021improving}
Zheng, L.; Tang, M.; Chen, Y.; Zhu, G.; Wang, J.; and Lu, H. 2021.
\newblock Improving multiple object tracking with single object tracking.
\newblock In \emph{Proceedings of the IEEE/CVF Conference on Computer Vision
  and Pattern Recognition}, 2453--2462.

\bibitem[{Zhou and Yuan(2016)}]{zhou2016learning}
Zhou, C.; and Yuan, J. 2016.
\newblock Learning to integrate occlusion-specific detectors for heavily
  occluded pedestrian detection.
\newblock In \emph{Asian Conference on Computer Vision}, 305--320. Springer.

\bibitem[{Zhou and Yuan(2018)}]{zhou2018bi}
Zhou, C.; and Yuan, J. 2018.
\newblock Bi-box regression for pedestrian detection and occlusion estimation.
\newblock In \emph{Proceedings of the European Conference on Computer Vision
  (ECCV)}, 135--151.

\bibitem[{Zhou, Wang, and Kr{\"a}henb{\"u}hl(2019)}]{zhou2019objects}
Zhou, X.; Wang, D.; and Kr{\"a}henb{\"u}hl, P. 2019.
\newblock Objects as points.
\newblock \emph{arXiv preprint arXiv:1904.07850}.

\end{thebibliography}

\end{document}